\documentclass[acmsmall]{acmart} 

\newcommand{\hdh}{HDHumans}
\setcopyright{rightsretained} 
\acmJournal{PACMCGIT}
\acmYear{2023} \acmVolume{6} \acmNumber{2} \acmArticle{} \acmMonth{8} \acmPrice{}\acmDOI{10.1145/3606927}

\begin{document}

\title{HDHumans: A Hybrid Approach for High-fidelity Digital Humans}

\author{Marc Habermann}
\affiliation{%
	\institution{Max Planck Institute for Informatics}
	\country{Germany}
}
\email{mhaberma@mpi-inf.mpg.de}
\author{Lingjie Liu}
\affiliation{%
	\institution{Max Planck Institute for Informatics}\country{Germany}
}
\email{lliu@mpi-inf.mpg.de}
\author{Weipeng Xu}
\affiliation{%
	\institution{Meta Reality Labs}\country{United States}
}
\email{xuweipeng@meta.com}
\author{Gerard Pons-Moll}
\affiliation{%
	\institution{University of Tuebingen}
	\country{Germany}
}
\email{gerard.pons-moll@uni-tuebingen.de}
\author{Michael Zollhoefer}
\affiliation{%
	\institution{Meta Reality Labs}\country{United States}
}
\email{zollhoefer@meta.com}
\author{Christian Theobalt}
\affiliation{\institution{Max Planck Institute for Informatics}\country{Germany}
 }
 \email{theobalt@mpi-inf.mpg.de}


%
%
\begin{teaserfigure}
	\centering
\includegraphics[width=\linewidth]{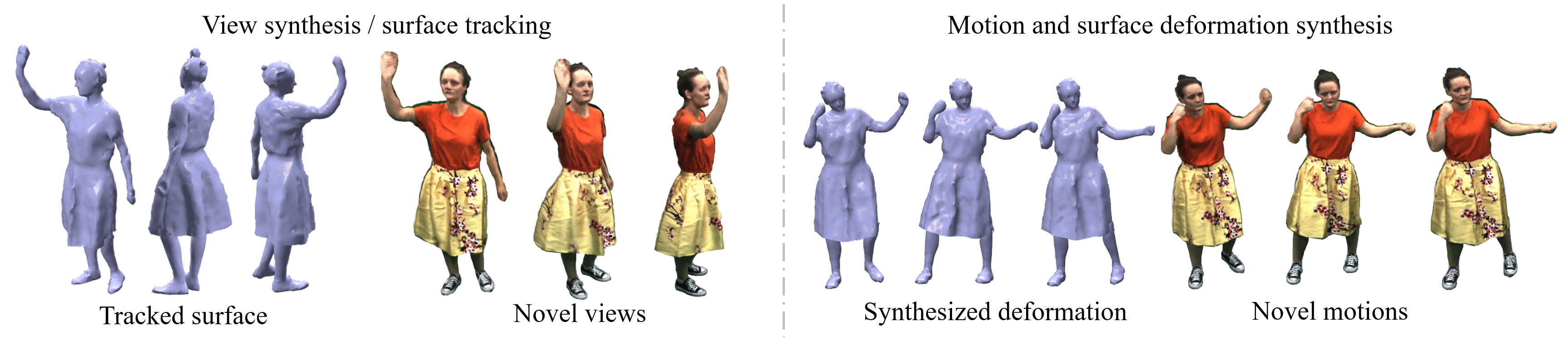}
	\caption
	{
		We propose a method for photo-realistic human synthesis given an arbitrary camera pose and a potentially unseen skeletal motion.
		Our method also handles loose types of clothing such as skirts, since we jointly learn the dense and space-time coherent deforming geometry of the human surface (including the dynamic clothing) along with a neural radiance field. 
	}
	\label{fig:futureTeas}
\end{teaserfigure}
%
%

\begin{abstract}
Photo-real digital human avatars are of enormous importance in graphics, as they enable immersive communication over the globe, improve gaming and entertainment experiences, and can be particularly beneficial for AR and VR settings.
However, current avatar generation approaches either fall short in high-fidelity novel view synthesis, generalization to novel motions, reproduction of loose clothing, or they cannot render characters at the high resolution offered by modern displays. 
To this end, we propose \hdh, which is the first method for HD human character synthesis that jointly produces an accurate and temporally coherent 3D deforming surface and highly photo-realistic images of arbitrary novel views and of motions not seen at training time.
At the technical core, our method tightly integrates a classical deforming character template with neural radiance fields (NeRF).
Our method is carefully designed to achieve a synergy between classical surface deformation and a NeRF.
First, the template guides the NeRF, which allows synthesizing novel views of a highly dynamic and articulated character and even enables the synthesis of novel motions.
Second, we also leverage the dense pointclouds resulting from the NeRF to further improve the deforming surface via 3D-to-3D supervision.
We outperform the state of the art quantitatively and qualitatively in terms of synthesis quality and resolution, as well as the quality of 3D surface reconstruction.
\end{abstract}





\maketitle


%
%
\section{Introduction} \label{sec:ongoing}
%
%
\par 
Photo-realistic synthesis of digital humans is a very important research topic in graphics and computer vision. Specially, with the recent developments of VR and AR headsets, it has become even more important, since photo-real human avatars can be used to populate virtual or augment real scenes.
The classical approach to achieve this goal would be the manual creation of human avatars by means of 3D modeling including meshing, texturing, designing material properties, and many more manual steps.
However, this process is not only tedious and time-consuming, but it also requires expert knowledge, preventing these techniques from being adopted by non-expert users.
A promising alternative is to create such digital human avatars from video captures of real humans.
The goal of our approach is to create controllable and highly photo-realistic characters at high resolution solely from multi-view video.
%
%
\par 
This idea was already subject of previous research works that can be broadly categorized based on the employed representation.
Some approaches explicitly model the human's surface as a mesh and employ texture retrieval techniques~\cite{casas14, Xu:SIGGRPAH:2011} or deep learning~\cite{habermann21} to generate realistic appearance effects.
However, the synthesis quality is still limited and the recovered surface deformations are of insufficient quality because they are driven purely by image-based supervision.
Other works solely synthesize humans in image space~\cite{liu19,chan2019dance, liu20}.
These approaches however suffer from 3D inconsistency when changing viewpoint.
Recently, first attempts have also been proposed to combine a neural radiance field with a human body model~\cite{liu2021neural,chen2021animatable,xu2021hnerf,peng2021animatable,peng2020neural}.
These works have demonstrated that a classical mesh-based surface can guide a neural radiance field (NeRF)~\cite{mildenhall2020nerf} for image synthesis.
However, since they rely on a human body model or skeleton representation, they do not model the underlying deforming surface well.
In consequence, they only work for subjects wearing tight clothing.
In stark contrast, we for the first time demonstrate how a NeRF can be conditioned on a \textit{densely deforming} template and we even show that improvements can be achieved in the other direction as well where the NeRF is guiding the mesh deformation.
%
%
\par 
In contrast to prior work, we propose a tightly coupled hybrid representation consisting of a classical deforming surface mesh and a neural radiance field defined in a thin shell around the surface.
On the one hand, the surface mesh guides the learning of the neural radiance field, enables the method to handle large motions and loose clothing, and leads to a more efficient sampling strategy along the camera rays. 
On the other hand, the radiance field achieves a higher synthesis quality than pure surface-based approaches, produces explicit 3D constraints for better supervision of explicit surface deformation networks, and helps in overcoming local minima due to the local nature of color gradients in image space.
This tight coupling between explicit surface deformation and neural radiance fields creates a two-way synergy between both representations.
We are able to jointly capture the detailed underlying deforming surface of the clothed human and also employ this surface to drive a neural radiance field, which captures high-frequency detail and texture.
More precisely, our method takes skeletal motion as input and predicts a motion-dependent deforming surface as well as a motion- and view-dependent neural radiance field that is parameterized in a thin shell around the surface.
In this way, the deforming surface acts as an initializer for the sampling and the feature accumulation of the neural radiance field making it significantly ($6$ times) more efficient and, thus, enables training on 4K multi-view videos.
The deforming surface mesh and the neural radiance field are tightly coupled during training such that the mesh drives the neural radiance field making it efficient and robust to dynamic changes.
Furthermore, not only the neural radiance field is improved based on the tracked surface mesh, but it can also be used to refine the surface mesh, since the neural radiance field drives the mesh towards reconstructing finer-scale detail, such as cloth wrinkles, which are difficult to capture with image-based supervision alone.
Thus, a two-way synergy between the employed classical and neural scene representation is created that leads to significantly improved fidelity.
Compared to previous work, our approach not only reconstructs deforming surface geometry of higher quality, but also renders human images at much higher fidelity (see Figure~\ref{fig:futureTeas}).
%
%
In summary, our technical contributions are:
\begin{itemize}
	\item A novel approach for high-fidelity character synthesis that enables novel view and motion synthesis at a very high resolution, which cannot be achieved by previous work.
	\item A synergistic integration of a classical mesh-based and a neural scene representation for virtual humans that produces higher quality geometry, motion, and appearance than any of the two components in isolation.
\end{itemize}
To the best of our knowledge, this is the first approach that tightly couples a \textit{deforming explicit mesh} and a NeRF enablings photo-realistic rendering of neural humans \textit{wearing loose clothing}.
%
%
%
\section{Related Work} \label{sec:relatedwork}
%
%
\par \textbf{Mesh-based synthesis.}
Photo-realistic image synthesis of controllable characters is challenging due to the difficulty in capturing or predicting high-quality pose-dependent geometry deformation and appearance.
Some works~\cite{carranza03,Li:2014,zitnick04,collet15,hilsmann20} focus on free-viewpoint replay of the captured human performance sequence.
Other works~\cite{Xu:SIGGRPAH:2011,casas14,Volino2014} aim at the more challenging task of photo-realistic free-viewpoint synthesis for new body poses.
However, their method needs several seconds to generate a single frame.
\citet{casas14} and \citet{Volino2014} accelerate the image synthesis process with a temporally coherent layered representation of appearance in texture space.
These classical methods struggle with producing high-quality results due to the coarse geometric proxy, and have limited generalizability to new poses and viewpoints, which are very different from those in the database.
To improve the synthesis quality and generalizability, \citet{habermann21} proposes a method for learning a 3D virtual character model with pose-dependent geometry deformations and pose- and view-dependent textures in a weakly supervised way from multi-view videos.
While great improvements have been made, some fine-scale details are missing in the results, because of the difficulty in the optimization of deforming polygon meshes with only images as supervision.
In this work, we observed that deforming implicit fields is more flexible (e.g., no need of using regularization terms to keep the mesh topology), thus leading to more stable and efficient training.
However, the rendering of implicit fields is time-consuming, and editing implicit representations is much more difficult than editing explicit representations, e.g., meshes.
Hence, our method unifies the implicit fields and explicit polygon meshes joining the advantages from both worlds.
%
%
\par \textbf{Image-based synthesis.}
GANs have achieved great progress in image synthesis in recent years. 
To close the gap between the rendering of a coarse geometric proxy and realistic renderings, many works formulate the mapping from the coarse rendering to a photo-realistic rendering as an image-to-image translation problem.
These works take the renderings of a skeleton~\cite{chan2019dance,Pumarola_2018_CVPR,zhu2019progressive,kappel2020high-fidelity,Shysheya_2019_CVPR,li2019dense}, a dense mesh~\cite{liu19,wang2018vid2vid,liu20,Sarkar2020,Neverova2018,Grigorev2019CoordinateBasedTI,lwb2019, Raj_ANR, SMPLpix:WACV:2020}, or a joint position heatmap~\cite{MaSJSTV2017,Lischinski2018,Ma18} as the input to image-to-image translation and output realistic renderings.
While these methods can produce high-quality images from a single view, they are not able to synthesize view-consistent videos when changing camera viewpoints. 
In contrast, our method directly optimizes the geometry deformations and appearance in 3D space, so it is able to produce temporally- and view-consistent photo-realistic animations of characters.
%
%
\par \textbf{Volume-based and hybrid approaches.}
Recently, some methods have demonstrated impressive results on novel view synthesis of static scenes by using neural implicit fields~\cite{sitzmann2019scene, mildenhall2020nerf, Oechsle2021ICCV, wang2021neus, yariv2021volume, yariv2020multiview, DVR} or hybrid representations~\cite{liu2020neural, hedman2021snerg, Reiser2021ICCV, yu2021plenoctrees, devries2021unconstrained} as scene representations. 
Great efforts have been made to extend neural representations to dynamic scenes.
Neural Volumes~\cite{lombardi2019neural} and its follow-up work~\cite{wang2020learning} use an encoder-decoder network to learn a mapping from reference images to 3D volumes for each frame of the scene, followed by a volume rendering technique to render the scene.
Several works extend the NeRF~\cite{mildenhall2020nerf} to dynamic scene modeling with a dedicated deformation network~\cite{tretschk2020nonrigid,park2020nerfies, pumarola2020dnerf,park2021hypernerf}, scene flow fields~\cite{Li2020NeuralSF}, or space-time neural irradiance fields~\cite{xian2020spacetime}.
Many works focus on human character modeling. 
\citet{peng2020neural} and \citet{kwon2021neural} assign latent features on the vertices of the SMPL model and use them as anchors to link different frames.
\citet{lombardi2021mixture} introduce a mixture of volume primitives for the efficient rendering of human actors.
These methods can only playback a dynamic scene from novel views but are not able to generate images for novel poses. 
To address this issue, several methods propose articulated implicit representations for human characters. 
A-NeRF~\cite{su2021anerf} proposes an articulated NeRF representation based on a human skeleton for human pose refinement. 
Recent works~\cite{liu2021neural,peng2021animatable,chen2021animatable,xu2021hnerf, 2021narf, zhang2021stnerf, NNA, li2022tava, ARAH:ECCV:2022} present a deformable NeRF representation, which unwarps different poses to a shared canonical space with inverse kinematic transformations and residual deformations.
Moreover, HumanNeRF~\cite{weng2022humannerf} has shown view-synthesis for human characters given only a monocular RGB video for training.
Most of these works cannot synthesize pose-dependent dynamic appearance, are not applicable to large-scale datasets that include severe pose variations, and have limited generalizability to new poses. 
The most related work to our proposed method is Neural Actor~\cite{liu2021neural}, which uses a texture map as a structure-aware local pose representation to infer dynamic deformation and appearance.
In contrast to our method, they only use a human body model as a mesh proxy, and thus cannot model characters in loose clothes. 
Furthermore, they only employ the mesh proxy to guide the warping of the NeRF but do not optimize the mesh. 
In consequence, this method cannot extract high-quality surface geometry. 
Further, since the mesh proxy is not very close to the actual surface, it still needs to sample many points around the surface, which prevents training on 4K resolution. 
Instead, we infer the dense deformation of a template that is assumed to be given, which is more efficient and enables the tracking of loose clothing.
More importantly, our recovered NeRF even further refines the template deformations.
%
%
%
\section{Method} \label{sec:overviewFuture}
\begin{figure*}[t]
	\begin{center}
		\includegraphics[width=\linewidth]{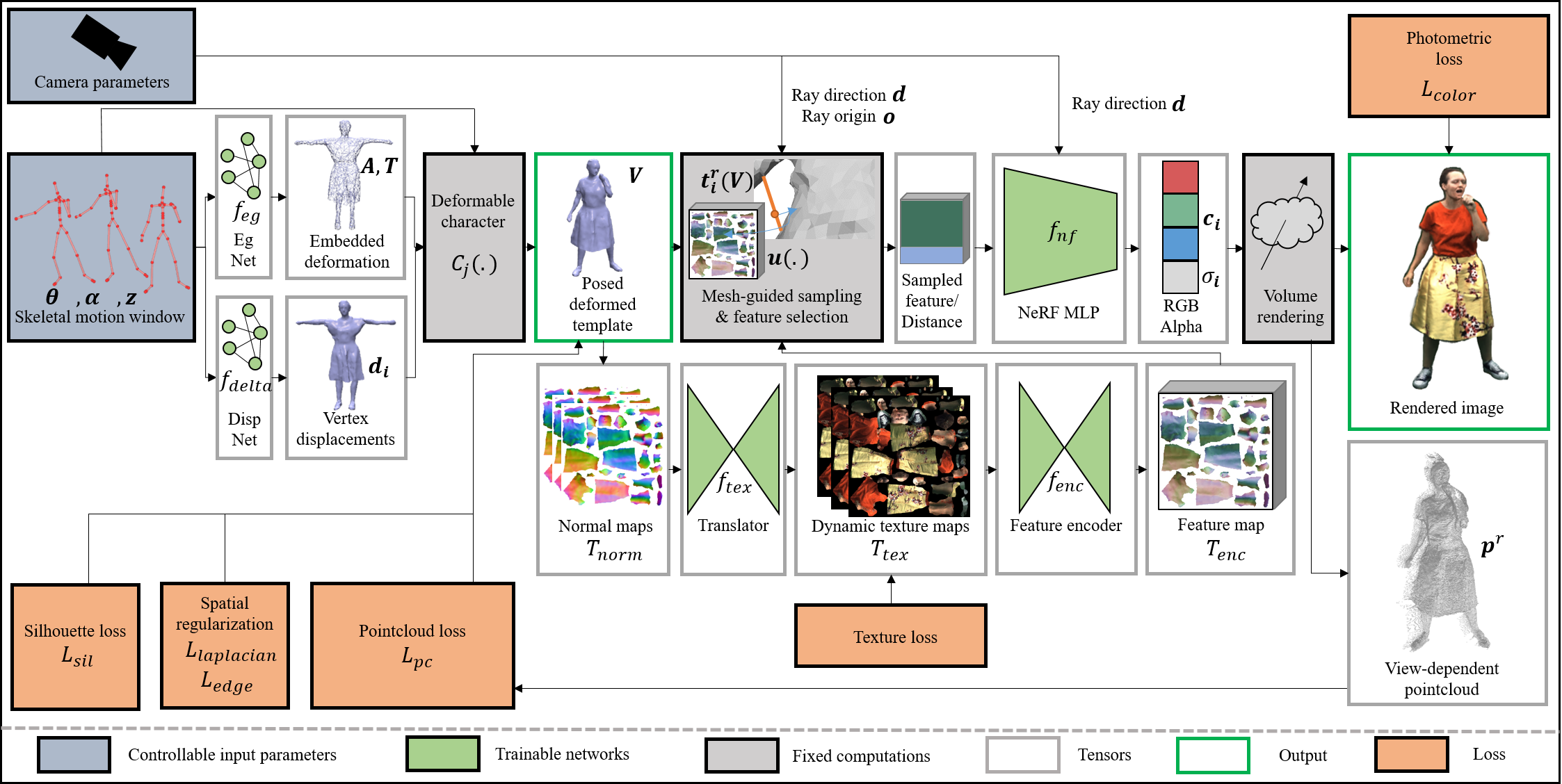}
	\end{center}
	\caption
	{
		Overview of the proposed approach.
		Our method takes as input a skeletal motion of the actor and predicts high-quality appearance as well as space-time coherent and deforming geometry.
	}
	\label{fig:futureOverview}
\end{figure*}

The goal of our approach is to learn a unified representation of a dynamic human from multi-view video, which on the one hand allows to synthesize motion-dependent  deforming geometry and on the other hand also enables photo-real synthesis of images displaying the human under novel viewpoints and novel motions.
To this end, we propose an end-to-end approach, which solely takes a skeletal motion and a camera pose as input and outputs a posed and deformed mesh as well as the respective photo-real rendering of the human.
Figure~\ref{fig:futureOverview} shows an overview of the proposed method.
In the following, some fundamentals are provided (Section~\ref{sec:background}).
Then, we introduce our mesh-guided neural radiance field, which allows synthesizing a dynamic performance of the actor from novel views and for unseen motions (Section~\ref{sec:meshguided}).
This proposed mesh guidance assumes a highly detailed, accurately tracked, and space-time coherent surface of the human actor.
We however found that previous weakly-supervised performance capture approaches~\cite{habermann20, habermann21} struggle with capturing high fidelity geometry.
At the same time, volume-based surface representations~\cite{mildenhall2020nerf} seem to recover such geometric details when visualizing their view-dependent pointclouds, but they lack space-time coherence, which is essential for the proposed mesh guidance.
To overcome this limitation, we propose a NeRF-guided point cloud loss, which further improves the motion-dependent and deformable human mesh model (Section~\ref{sec:nerfguided}).
%
%
%
%
%
%
%
\par \textbf{Data assumptions.}
For each actor, we employ $C$ calibrated and synchronized cameras to collect a segmented multi-view video of the person performing various types of motions.
The skeletal motion that is input to our method is recovered using a markerless motion capture software~\cite{captury}.
Finally, we acquire a static textured mesh template of the person using a scanner~\cite{treedys} that is manually rigged to the skeleton.
Note that our approach does not assume any 4D geometry in terms of per-frame scans or pointclouds as input.
%
%
%
\subsection{Human Model and Neural Radiance Fields} \label{sec:background}
%
%
%
\subsubsection{Deformable Human Mesh Model} \label{sec:human_mesh_model}
%
%
We aim at having a deformation model of the human body and clothing, which only depends on the skeletal motion $\mathcal{M}=\{ (\theta_{t-T},\boldsymbol{\alpha}_{t-T},\mathbf{z}_{t-T}),...,(\theta_{t},\boldsymbol{\alpha}_{t},\mathbf{z}_{t}) \}$ and
deforms the person-specific template such that motion-dependent clothing and body deformations can be modeled, e.g. the swinging of a skirt induced by the motion of the hips.
Here, $\theta_{t}\in \mathbb{R}^{57}$, $\boldsymbol{\alpha}_{t}\in \mathbb{R}^3$, and $\mathbf{z}_{t}\in \mathbb{R}^3$ refer to the skeletal joint angles, the root rotation and translation, respectively.
$(\cdot)_t$ refers to the $t$th frame of the video.
In practice, the time window is set to 3 ($T=2$) and for the current character pose $(\theta_{t},\boldsymbol{\alpha}_{t},\mathbf{z}_{t})$ we drop the frame subscript for readability in the following and assume the motion window is fixed.
We leverage the character representation of Habermann et al.~\shortcite{habermann21}
%
%
\begin{equation}\label{eq:characterrep}
\mathcal{C}_j(\boldsymbol{\theta}, \boldsymbol{\alpha}, \mathbf{z}, \mathbf{A}, \mathbf{T}, \mathbf{d}_j) = \mathbf{v}_j \mathrm{,}
\end{equation}
%
%
which takes the current skeletal pose, embedded deformation parameters $\mathbf{A}, \mathbf{T} \in \mathbb{R}^{K \times 3}$, where $K$ denotes the number of graph nodes of an underlying graph, and the per vertex displacement $\mathbf{d}_j \in \mathbb{R}^3$ for vertex $j$ as input.
It returns the posed \textit{and} deformed vertex $\mathbf{v}_j$.
%
%
\par
In order to model the skeletal motion-dependent template deformation, embedded deformations~\cite{sumner07, sorkine07} are learned by a motion-conditioned graph convolutional network
%
%
\begin{equation} \label{eq:f_eg}
    f_{\mathrm{eg}}(e(\mathcal{M})) = \mathbf{A}, \mathbf{T} 
\end{equation}
%
%
where $e(\cdot)$ is their proposed skeletal motion-to-graph encoding.
To capture dynamic details beyond the resolution of the embedded graph, a second network
%
%
\begin{equation} \label{eq:f_delta}
f_{\mathrm{delta}}(d(\mathcal{M}))_j= \mathbf{d}_j
\end{equation}
%
%
learns the additional per-vertex displacements as a function of the skeletal motion.
Here, $d(\cdot)$ is their proposed motion-to-vertex encoding.
Now, inserting both networks into Equation~\ref{eq:characterrep} leads to the generative character model
%
%
\begin{equation}\label{eq:fineCharacter}
\mathcal{C}_j(\boldsymbol{\theta}, \boldsymbol{\alpha}, \mathbf{z}, f_{\mathrm{eg}}(e(\mathcal{M})), f_{\mathrm{delta}}(d(\mathcal{M}))_j) = \mathbf{v}_{j}
\end{equation}
%
%
which is solely parameterized by the skeletal pose $(\theta, \boldsymbol{\alpha}, \mathbf{z})$ and the network outputs that are conditioned on the skeletal motion $\mathcal{M}$.
The entire posed and deformed mesh can be derived by stacking the individual vertices into a matrix $\mathbf{V} \in \mathbb{R}^{N \times 3}$ where $N$ denotes the number of template vertices.
%
%
Interestingly, the deformation networks can be trained purely from image data using a multi-view silhouette loss $\mathcal{L}_\mathrm{sil}$ and a dense rendering loss $\mathcal{L}_\mathrm{chroma}$, as well as some spatial regularizers $\mathcal{L}_\mathrm{spatial}$.
We follow their proposed training procedure to obtain pre-trained deformation networks.
%
%
%
\subsubsection{Neural Radiance Fields} \label{sec:nerf}
A neural radiance field (NeRF)~\cite{mildenhall2020nerf} is a deep, volumetric scene representation of a static scene, which enables photo-realistic novel view synthesis.
In detail, for rendering an image, each pixel is represented by a camera ray that has a normalized direction $\mathbf{d} \in \mathbb{R}^3$ and an origin $\mathbf{o} \in \mathbb{R}^3$.
Then, $i \in \{0, ..., K\}$ samples along the ray at positions 
%
%
\begin{equation} \label{eq:nerf_sampling_ori}
\mathbf{x}_i = \mathbf{o} + t_i \mathbf{d}~, 
\end{equation} 
%
%
where $\mathbf{x}_i \in \mathbb{R}^3$ and $t_i$ is the depth along the ray, are drawn and fed into a Multi-layer Perceptron (MLP)
%
%
\begin{equation} \label{eq:nerfEval}
f_{\mathrm{nf}}(\gamma(\mathbf{x}_i), \gamma(\mathbf{d}))= (\mathbf{c}_i, \sigma_i)
\end{equation}
%
%
which takes the positional encoding $\gamma(\cdot)$ of $\mathbf{x}_i$ and $\mathbf{d}$ as input.
Then, the network predicts a color $\mathbf{c}_i \in \mathbb{R}^3$ and a density value $\sigma_i \in \mathbb{R}$.
\par
To obtain the final pixel color, the individual colors and densities are accumulated using volume rendering~\cite{levoyVolumeRendering} according to
%
%
\begin{equation} \label{eq:volumerender}
\tilde{\mathbf{c}} = \sum_{i=0}^K T_i \alpha_i \mathbf{c}_i, \quad T_i = \prod_{j=0}^{i-1} (1-\alpha_j), \quad  \alpha_i = 1 - e^{-\delta_i \sigma_i}
\end{equation} 
%
%
where $\delta_i$ is the Euclidean distance between $\mathbf{x}_{i+1}$ and $\mathbf{x}_{i}$.
%
%
%
\subsubsection{Discussion} \label{sec:background_discussion}
The main advantage of the proposed geometry-based character representation~\cite{habermann21} is that it can represent dynamically moving humans and that the reconstructed and synthesized geometry matches image silhouettes well and it shows some plausible coarse deformations.
However, there is still a gap in terms of surface accuracy and the approach suffers from baked-in geometric details originating from the scanned template mesh, which remain almost rigid throughout the deformation.
We suspect that this comes from the purely image-based supervision strategy, which prevents the template from being deformed more drastically with respect to the scan.
While they enable the synthesis of loose clothing such as skirts, we also found that the learned dynamic texture and the final appearance in image space cannot reach the quality of volume-based neural rendering approaches for static scenes.
%
%
\par
NeRF~\cite{mildenhall2020nerf} has shown state-of-the-art synthesis quality on \textit{static} scenes.
Interestingly, when training a NeRF on humans, the recovered depth maps show detailed wrinkle patterns despite some noise and outliers.
However, long compute time and the fact that the original NeRF formulation is limited to static scenes prevent it from being directly used on long dynamic scenes (around $20{,}000$ frames), which we are targeting.
Recently, human-specific follow-up works~\cite{liu2021neural,chen2021animatable,xu2021hnerf,peng2021animatable,peng2020neural} have been introduced.
However, they usually rely on a human body model and do not account for non-rigid surface deformations, such as the dynamic movement of clothing.
As a consequence, these works are limited to types of apparel that tightly align with the human body while loose clothes such as skirts are beyond their reach.
%
%
\par
In the following, we address these problems of 1) limited geometric deformations caused by the purely image-based supervision, 2) limited synthesis quality of geometry-based representations, and 3) the limited types of apparel that can be synthesized by recent NeRF-based approaches.
To this end, we propose a novel and non-trivial combination of a deformable mesh-based representation and a neural radiance field and show that one can overcome the above limitations by tightly coupling those two representations.
%
%
\subsection{Mesh-guided Neural Radiance Fields} \label{sec:meshguided}
First, we establish a tight connection between the deforming human mesh model and a surrounding neural radiance field.
Here, we assume the posed and deformed vertex positions $\mathbf{V}$ are given by our pre-trained character model (Equation~\ref{eq:fineCharacter}).
%
%
%
\subsubsection{Motion-dependent Neural Texture} \label{sec:motion_texture}
When thinking about defining motion-dependent features on the mesh surface, there usually is the problem of an one-to-many mapping~\cite{bagautdinov21, liu2021neural}, since (almost) similar motion can lead to various different images, i.e., wrinkle patterns on the clothing.
This is due to the fact that the state of clothing does not only depend on the last few poses of the actor, but there exist also other factors.
Examples are the initial state of clothing when the performance starts, external forces such as wind, and second order dynamics.
On the one hand, reliably modeling all of these is intractable and on the other hand ignoring them leads to a blurred appearance~\cite{habermann21}.
%
%
Similar to Liu et al.~\cite{liu2021neural}, we encode the actor's pose $(\theta_{t},\boldsymbol{\alpha}_{t},\mathbf{z}_{t})$ of frame $t$ in the form of a normal map $\mathcal{T}_{\mathrm{norm},t} \in \mathbb{R}^{1024 \times 1024 \times 3}$, which is denoted by the function $m(\theta_{t},\boldsymbol{\alpha}_{t},\mathbf{z}_{t})= \mathcal{T}_{\mathrm{norm},t}$.
However, different to them we use the posed \textit{and deformed} geometry for creating the normal maps.
Thus, higher frequency geometric details are explicitly represented in the normal maps.
Then a texture-to-texture translation network~\cite{wang2018vid2vid}
%
%
\begin{equation} \label{eq:f_tex}
      f_\mathrm{tex} (m(\theta_{t},\boldsymbol{\alpha}_{t},\mathbf{z}_{t})) = f_\mathrm{tex} (\mathcal{T}_{\mathrm{norm},t} ) = \mathcal{T}_{\mathrm{tex}, t}
\end{equation}
%
%
converts them into dynamic texture maps $\mathcal{T}_{\mathrm{tex},t} \in \mathbb{R}^{1024 \times 1024 \times 3}$, which contain realistic cloth wrinkle patterns.
For creating the training pairs, the posed normal maps can be trivially computed from the posed and deformed mesh.
For generating the ground truth texture maps $\mathcal{T}_{\mathrm{tex, gt},t} \in \mathbb{R}^{1024 \times 1024 \times 3}$, we leverage the multi-view texture stitching approach proposed by Alldieck et al.~\cite{alldieck18a}.
Simply using an $\ell_1$-loss for $\mathcal{L}_\mathrm{tex}$ between $\mathcal{T}_{\mathrm{tex, gt,t}}$ and $\mathcal{T}_{\mathrm{tex},t}$ would still lead to blurry results for the above mentioned reasons.
Thus, we employ a discriminator loss as proposed in the vid2vid architecture~\cite{wang2018vid2vid}.
This greatly reduces the problem of the one-to-many mapping.
%
%
\par 
Finally, we have a UNet-based feature encoder~\cite{pix2pix2017}, 
%
%
\begin{equation} \label{eq:f_enc}
    f_\mathrm{enc} (\mathcal{T}_{\mathrm{tex},t-T}, ..., \mathcal{T}_{\mathrm{tex},t} ) = \mathcal{T}_{\mathrm{enc}, t}
\end{equation}
%
%
which takes the generated textures $\mathcal{T}_{\mathrm{tex}, t'}$ of the motion window $t' \in \{t-T, ...,t\}$ by concatenating them along the last channel resulting in a texture tensor $\mathcal{T}_{\mathrm{enc}, t} \in \mathbb{R}^{1024 \times 1024 \times 3(T+1)}$.
The output of this is a feature texture $\mathcal{T}_{\mathrm{enc}, t} \in \mathbb{R}^{1024 \times 1024 \times 32}$ that will be later used as an input to the NeRF.
Previous work~\cite{liu2021neural} showed that encoding the texture into a feature space rather than directly using the generated texture as a conditioning input to the NeRF improves the synthesis quality. 
Thus, we follow this design choice, however, we choose a UNet-based~\cite{pix2pix2017} encoder rather than a ResNet-based~\cite{he16} architecture.
This allows us to efficiently encode a higher resolution feature map ($1024 \times 1024$ vs. $512 \times 512$~\cite{liu2021neural}).
Further, since we predict appearance from motion rather than from a single pose~\cite{liu2021neural}, we also concatenate the per-pose textures $\mathcal{T}_{\mathrm{tex}, t}$ in the feature channel.
%
%
\subsubsection{Geometry-guided Sampling} \label{sec:samplingFuture}
Next, we describe how the NeRF sampling process can be represented as a function of the deforming mesh $\mathbf{V}$ given a ray $r$ and therefore tightly connects the two representations.
Assuming the training camera and the ray $r$ are fixed, the $i$th sample $\mathbf{x}_i$ along the ray is originally defined by Equation~\ref{eq:nerf_sampling_ori} where $t_i$ is first sampled uniformly and later based on importance sampling.
As $r$ is fixed, $\mathbf{o}$ and $\mathbf{d}$ are pre-defined. 
However, we replace $t_i$ with the following geometry-dependent function 
\begin{equation} \label{eq:sampleNerf}
	t_i^{r} (\mathbf{V}) = 
	 (1 +\frac{i}{S}) (E(\Phi^{r}(\mathbf{V})) - t_\mathrm{mi} ) 
	+ \frac{i}{S}  (D(\Phi^{r}(\mathbf{V}))  + t_\mathrm{ma}) \mathrm{.}
\end{equation}
Here, $\Phi^{r}(\mathbf{V})$ is a depth renderer, which renders the depth map of the mesh with respect to the camera, and $r$ indicates the specific pixel that was rendered.
The function $D(\cdot)$ represents the dilation operator, which computes the maximum depth value in the depth map around the sampled location $r$.
Similarly, $E(\cdot)$ computes the eroded value or minimum value around the sampled location $r$.
We choose a kernel size of $9 \times 9$ for both operators.
The erosion and dilation ensure that the NeRF is also sampled on the foreground when the underlying mesh is erroneously not exactly overlaying the ground truth foreground mask.
Moreover, $t_\mathrm{mi}$ defines the volume that is sampled in front of the actual surface, and similarly $t_\mathrm{ma}$ defines the volume that is sampled behind the actual surface by ensuring that the distance between the rendered depth and the depth of the sample point does not exceed $t_\mathrm{mi}$ and $t_\mathrm{ma}$.
We set $t_\mathrm{mi} = t_\mathrm{ma}=4cm$ for all results.
Lastly, $S$ defines the number of samples along the ray.
When sampling $r$, only pixels that project onto the eroded/dilated depth maps are considered.
Otherwise, they are discarded during the NeRF evaluation described later.
%
%
\par 
This allows a more effective sampling of the neural radiance field since most samples are very close to the actual surface.
In practice, we only need a single NeRF MLP in contrast to \cite{mildenhall2020nerf}, which employ a coarse and a fine MLP.
Moreover, they draw 64 samples for the coarse MLP and 128 for the fine MLP.
Since our mesh provides accurate sampling guidance, we only require 32 samples for generating photo-realistic results.
This effectively means our mesh-guided sampling is 6 times more efficient than the baseline, which is especially important when training on 4K multi-view videos.
%
%
%
\subsubsection{Geometry-guided Motion Features} \label{sec:geometry_guided}
The other important property, which is missing in the original NeRF approach, is that it can only render a \emph{static} scene under novel views.
However, we aim at synthesizing novel views and performances of \emph{dynamic} scenes.
Fortunately, the posed and deformed template can also help to enable the synthesis of dynamic scenes using our motion-dependent feature texture $\mathcal{T}_{\mathrm{enc}}$.
The main idea is that the motion-dependent deep features attached to the mesh can be propagated to the 3D ray samples.
Then, instead of evaluating the NeRF MLP on global coordinates, we condition the MLP on a surface relative encoding using the signed distance and the texture features.
More specifically, Equation~\ref{eq:nerfEval} is modified as 
%
%
\begin{equation} \label{eq:new_nerf}
	f_{\mathrm{nf}}(
		u(
			\mathbf{V},
			\mathbf{x}_i, 
			\mathcal{T}_{\mathrm{enc}})
	,\gamma(d(\mathbf{V},\mathbf{x}_i))
	,\gamma(\mathbf{d}))= (\mathbf{c}_i, \sigma_i).
\end{equation}
%
%
Here, $u(\cdot)$ takes the mesh $\mathbf{V}$, the sample $\mathbf{x}_i$ along the ray, and the feature texture $\mathcal{T}_{\mathrm{enc}}$ as input and samples the feature texture at the UV coordinate of the closest point from $\mathbf{x}_i$ to the mesh resulting in a 32-dimensional feature.
$d(\cdot)$ computes the signed and normalized distance between the mesh and the sample point.
Here, points in the interior of the mesh have a negative sign, and the points outside the mesh have a positive sign. 
The term \emph{normalized} means that the actual distance is divided either by $t_\mathrm{mi}$ or $t_\mathrm{ma}$, depending on whether the sample point is inside or outside the mesh surface.
A positional encoding~\cite{mildenhall2020nerf} is then applied to the distance value using 10 frequencies.
Finally, the MLP also takes the positional encoding of the viewing direction using 4 frequencies.
Note that Equation~\ref{eq:new_nerf} only depends on the skeletal motion $\mathcal{M}$ (which then defines $\mathbf{V}$ and $\mathcal{T}_{\mathrm{enc}}$) and the camera pose (which then defines $\mathbf{x}_i$ and $\mathbf{d}$).
Thus, this reformulation allows the network to encode the dynamic motion of the actor and allows the NeRF to handle dynamically moving humans.
%
%
%
\subsubsection{Supervision} \label{sec:trainingFuture}
During training, we fix the pre-trained deformation networks $f_\mathrm{eg}$ and $f_\mathrm{delta}$ and the texture translation network $f_\mathrm{tex}$ and only train the feature encoder $f_\mathrm{enc}$ as well as the NeRF MLP $f_\mathrm{nf}$.
Assuming $C$ images of calibrated cameras for a fixed frame are given, a random foreground pixel $r$ from a random camera is chosen, which has the ground truth color $\mathbf{c}_{\mathrm{gt}}^{r}$.  
We employ an $\ell_1$ loss between the ground truth color and the estimated one 
\begin{equation}
	\mathcal{L}_{color}(\tilde{\mathbf{c}}^{r}) = \lVert \tilde{\mathbf{c}}^{r} -\mathbf{c}_{\mathrm{gt}}^{r}\rVert_1^1.
\end{equation}
%
%
%
\subsection{NeRF-guided Deformation Refinement} \label{sec:nerfguided}
\begin{figure}
	\begin{center}
		\includegraphics[width=\linewidth]{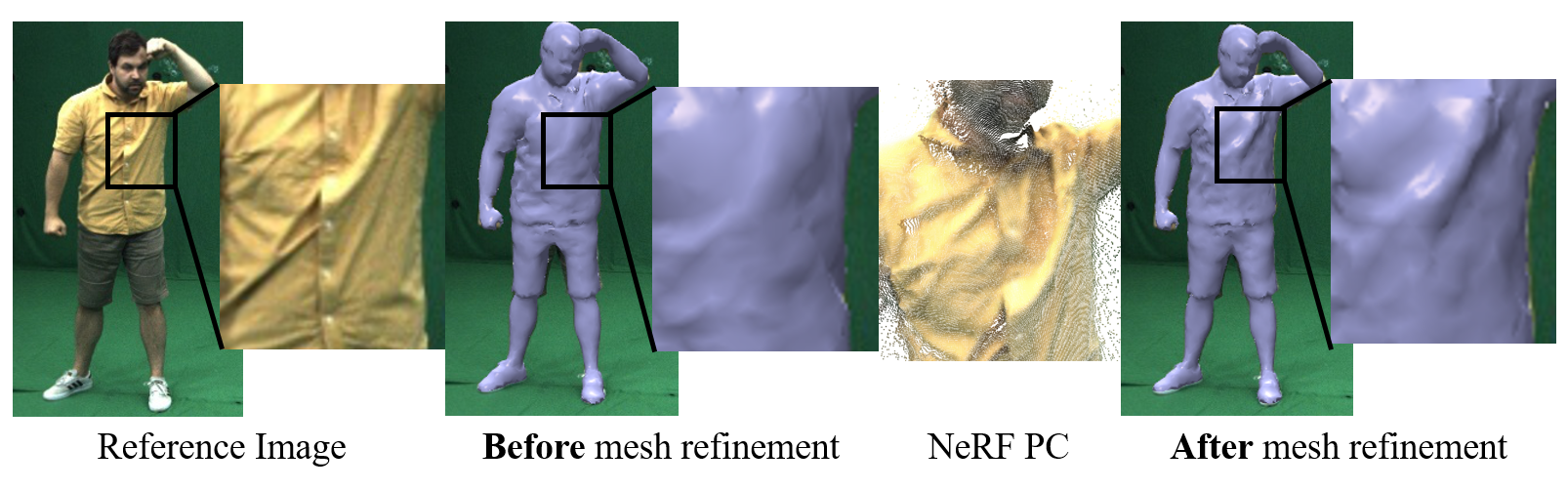}
	\end{center}
	\caption
	{
		Here, we visualize the influence of the NeRF-guided geomtry refinement using the proposed pointcloud loss.
		Note that the NeRF pointcloud contains much more geometric detail than the mesh \textit{before refinement.}
		Once we refined $f_\mathrm{delta}$ using the view-dependent pointcloud, these geometric details are also contained in the deformed mesh.
	}
	\label{fig:mesh_vs_nerf}
\end{figure}

So far, we have discussed how the NeRF representation can leverage the advantages of the underlying 3D template mesh.
However, the geometry can also be improved using the neural radiance field. 
The key observation is that a weakly supervised setup, as proposed by \cite{habermann21}, struggles with recovering the finer wrinkles on the clothing (see Figure~\ref{fig:mesh_vs_nerf}) due to the limited supervision.
For the silhouette loss $\mathcal{L}_\mathrm{sil}$, the main limiting factor is that it can at most recover details up to the visual hull, which is carved into the 3D volume by the multi-view foreground masks.
For the dense rendering loss $\mathcal{L}_\mathrm{chroma}$, there are three limitations: 1) the rendering loss is very sensitive to local minima as gradients of the input image are computed with finite differences on the ground truth image; 2) this loss struggles with deformations that are out of the camera plane, and 3) the rendering loss cannot account for shadows and view-dependent effects.
Fortunately, it can be observed that the per-view pointclouds that can be recovered from the proposed NeRF contain small-scale wrinkles (see Figure~\ref{fig:mesh_vs_nerf}). 
%
%
%
%
%
%
Thus, we supervise the template mesh by a 3D-to-3D constraint between the posed and deformed template and the per-view pointcloud, which has the advantage that no explicit per-frame multi-view stereo reconstruction is required.
%
%
%
\subsubsection{Poincloud Extraction} \label{sec:pointcloud_extraction}
The per-view pointcloud for a ray $r$ from any given camera of the current frame can be computed as 
%
%
\begin{equation}
	\label{eq:pointCloud}
	\mathbf{p}^{r} = 
	\mathbf{o}^{r}  + 
	\left(
		\sum_{i=0}^K   
			T_i^{r}  \alpha_i^{r}  t_i^{r}(\mathbf{V})
	\right)
	\mathbf{d}^{r}.
\end{equation}
%
%
Here, we use our mesh-guided NeRF representation (Equation~\ref{eq:new_nerf}) and volume rendering (Equation~\ref{eq:volumerender}) to acquire the density per ray sample $i$. 
This density is used to weight each (depth) sample along the ray and returns the average depth, which is then multiplied with the viewing direction $\mathbf{d}^{r}$.
Adding the camera origin $\mathbf{o}^{r}$ leads to the final point $\mathbf{p}^{r}$ in global 3D space.
We only sample rays for pixels where the dilated depth map is non-zero (foreground pixels).
For each frame and view, we sample $R=8192$ points, which we found is a good compromise between accuracy and training speed.
%
%
%
\subsubsection{Mesh Deformation Refinement} \label{sec:mesh_refinement}
%
%
Now, we further refine the mesh deformation network $f_\mathrm{delta}$ (while keeping the embedded deformation network $f_\mathrm{eg}$ fixed) using an additional loss
\begin{equation} \label{eq:chamferLossNerfBased}
\begin{aligned}
	\mathcal{L}_\mathrm{pc}(\mathbf{V})&=
	\sum_{j=0}^N
		\eta\left(
		min_{r \in \{0,...,R\}}
		\lVert 
			\mathbf{V}_j - \mathbf{p}^{r} 
		\rVert^2\right) \\
	&+
	\sum_{r=0}^R
		\eta\left(
		min_{j \in \{0,...,N\}}
		\lVert 
			\mathbf{p}^{r}  - \mathbf{V}_j
		\rVert^2\right)
\end{aligned}
\end{equation}
%
%
where $N$ is the number of template vertices and $\eta(\cdot)$ is a robust loss function that sets the value to zero when it exceeds a certain threshold $T=4cm$ to ensure robustness with respect to outliers.
Now, DeltaNet is refined with the losses
\begin{equation} \label{eq:loss_mesh}
   \mathcal{L}_\mathrm{mesh} = \mathcal{L}_\mathrm{pc}
   + \mathcal{L}_\mathrm{sil}
   + \mathcal{L}_\mathrm{laplacian}
   + \mathcal{L}_\mathrm{edge} \mathrm{.}
\end{equation}
Here, $\mathcal{L}_\mathrm{edge}$ is an isometry or edge length constraint that is imposed similar to the one proposed by \cite{habermann19}.
This constraint has the advantage that it allows local rotations in contrast to Laplacian regularization, which is important when trying to reproduce wrinkle patterns.
$\mathcal{L}_\mathrm{sil}$ and $\mathcal{L}_\mathrm{laplacian}$ are a multi-view silhouette loss and a Laplacian regularizer~\cite{habermann21}.
Figure~\ref{fig:mesh_vs_nerf} shows that the proposed NeRF pointcloud loss helps to recover finer wrinkles and to ensure that the deformed and posed template better matches the ground truth.
%
%
\par 
Once the mesh is refined, the whole process can be iterated.
We found that the refined geometry further improves the synthesis quality, which ultimately means that a synergy effect between the deformable mesh and the neural radiance field arises and both improve each other over each iteration till convergence is reached.
For more details concerning the training procedure and the implementation, we refer the reader to the supplemental document.

%
%
\section{Results} \label{sec:results}
%
%
\par \textbf{Dataset.}
We evaluate our proposed approach on the publicly available \textit{DynaCap} dataset~\cite{habermann21}, which contains 5 different actors.
Three actors wear loose clothes, i.e., two dresses and and one skirt.
The other two actors wear tighter clothing, i.e., short and long pants and long and short sleeves.
Each actor is performing a large variety of motions for the training and testing sequences.
Further, the motions in the test split significantly differ from the ones contained in the training set.
We follow the proposed train/test split of the dataset.
The original released dataset has an image resolution of $1285\times940$.
However, the authors of the dataset also provided us the full resolution videos ($4112\times3008$) for all sequences.
%
%
%
\subsection{Qualitative Results} \label{sec:qualitative}
%
%
%
\begin{figure*}
	\begin{center}
		\includegraphics[width=\linewidth]{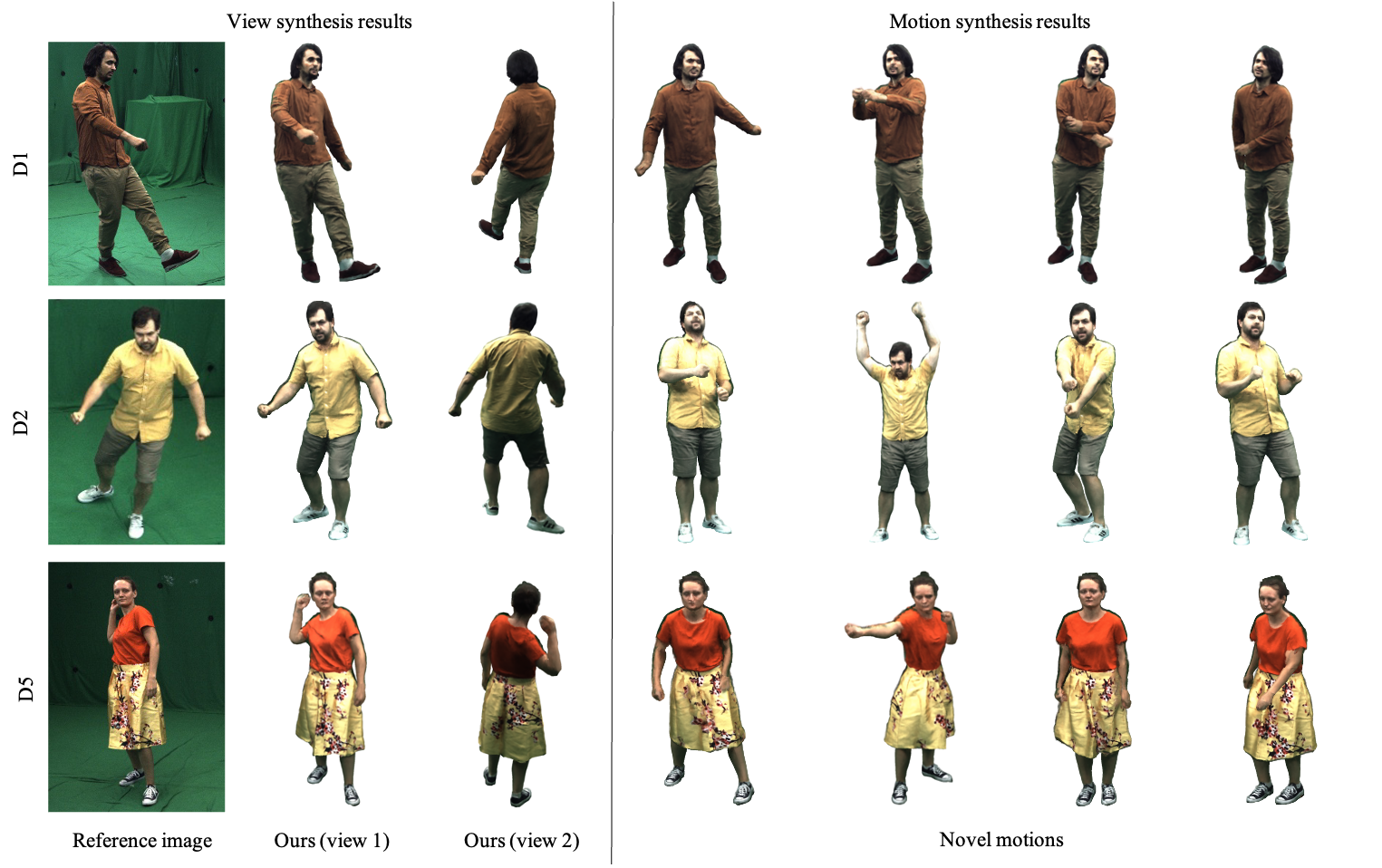}
	\end{center}
	\caption
	{
		Qualitative results.
        We show qualitative results for novel views and skeletal motions.
        Our method achieves a photo-realistic rendering quality and even individual clothing wrinkles appear sharp in the images.
	}
	\label{fig:qualitative}
\end{figure*}
%
%
%
First, we show qualitative results for the image synthesis quality of our approach in Figure~\ref{fig:qualitative} for all subjects in the dataset. 
We visualize novel view synthesis results for training motions (left column) as well as novel motion and view results (right column) and provide a reference image for each actor.
Note that in both modes, the results look highly photorealistic and even small clothing wrinkles can be realistically synthesized.
View-dependent appearance effects, such as view-dependent specular highlights on the skin of the actors, are also synthesized realistically. 
Notably, our method consistently achieves a high synthesis quality irrespective of the clothing type such that even loose clothing can be synthesized well.
\par
\begin{figure*}
	\begin{center}
		\includegraphics[width=\linewidth]{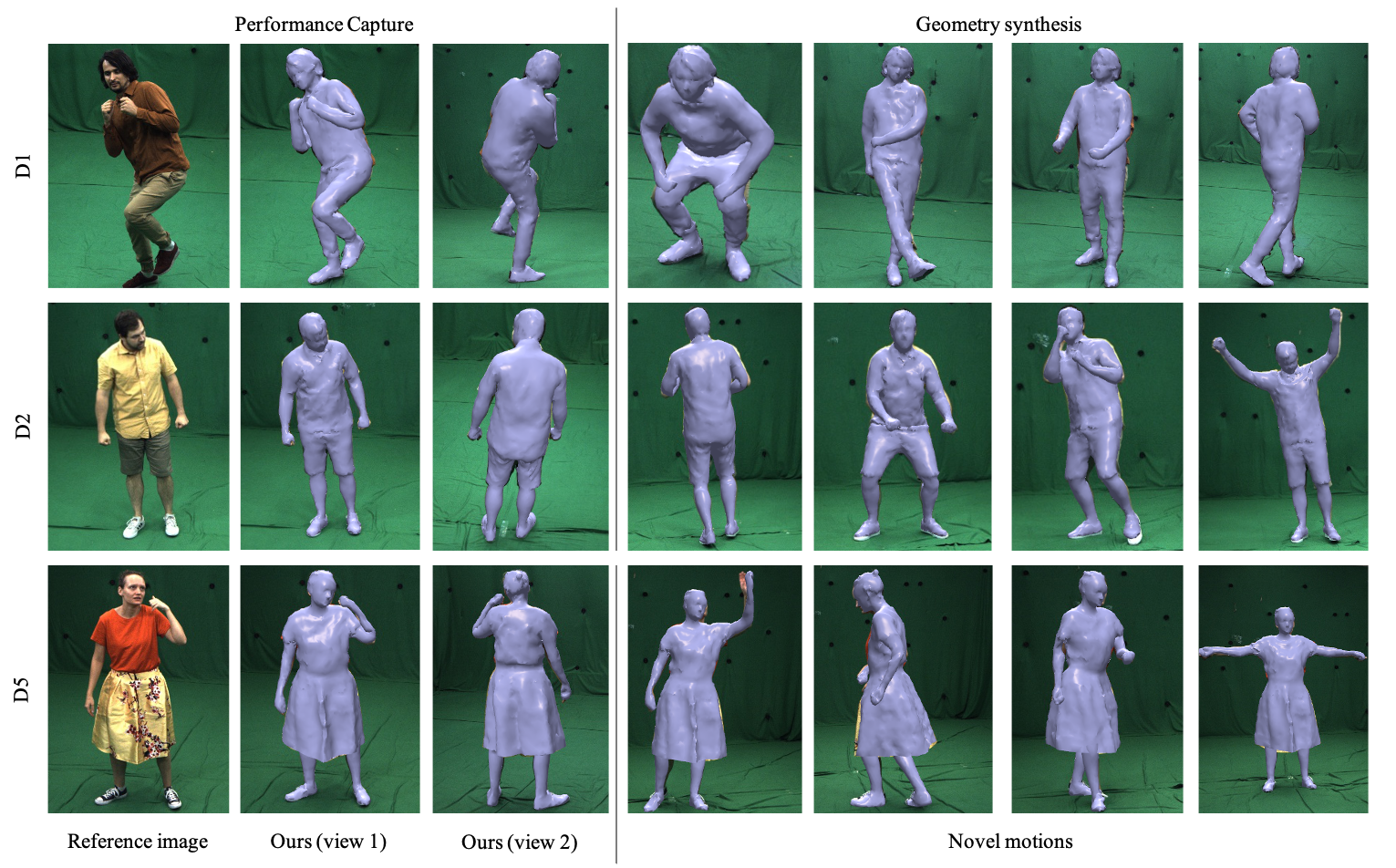}
	\end{center}
	\caption
	{
		Qualitative results showing our reconstructed/synthesized geometry on the \textit{DynaCap} dataset.
		Note that due to the novel NeRF-guided supervision of surface deformations, geometric details (such as clothing wrinkles) can be recovered nicely.
	}
	\label{fig:qualitative_rec}
\end{figure*}

We further show qualitative results of our space-time coherent geometry and synthesized motion-dependent deformations in Figure~\ref{fig:qualitative_rec}.
The 3D wrinkle patterns are nicely recovered in the geometry and the mesh also aligns well to the reference views.
\par 
Thus, our method is very versatile in the sense that it 1) faithfully reconstructs the geometry of the training sequence and 2) re-renders the training sequences from novel views.
Moreover, our method is also capable of 3) synthesizing motion-dependent surface deformations for unseen skeletal motions and 4) synthesizing photo-real images of the actor performing unseen skeletal motions.
%
%
%
%
%
%
For more qualitative results, we refer to our supplemental video.
%
%
\subsection{Comparisons to the State of the Art} \label{sec:comparison}
%
%
\subsubsection{Evaluation Sequence}
We compare to other methods on the challenging \textit{D2} sequence of the DynaCap dataset.
We do not evaluate on subjects with more loose clothing as except \citet{habermann21} and our method no previous work is able to track loose clothing.
Every metric is averaged across the entire sequence using every 10th frame. 
We hold out four cameras (with indices 7, 18, 27, and 40) for testing, which are uniformly sampled in space.
For quantitative evaluation, we also reconstruct pseudo ground truth geometry per frame using an off-the-shelf multi-view stereo reconstruction software~\cite{agisoftphotoscan}.
While the overall quality of the geometry is very high, slight reconstruction errors are unavoidable.
However, testing on real data is preferable as it is hard to faithfully simulate the complex and dynamic deformations and appearance effects that can be obvserved in this dataset.
For more qualitative comparisons, we also refer to our supplemental video.
%
%
\subsubsection{Previous Methods and Baselines}
We compare to Neural Actor (NA)~\cite{liu2021neural} and A-NeRF~\cite{su2021anerf}, which are also hybrid approaches in the sense that they attach a NeRF to a human body model~\cite{loper15} or skeleton.
We further compare to Neural Volumes (NV)~\cite{lombardi2019neural}, which is a neural volume rendering approach and Neural Body (NB)~\cite{peng2020neural}, which leverages structured latent codes that can be posed using an underlying skeleton structure.
We also compare to the surface and neural texture-based approach, Deep Dynamic Character (DDC)~\cite{habermann21}, which is the only related work that also tracks and synthesizes the underlying surface deformation.
Last, we compare to NHR~\cite{wu2020multi}, which uses a point-based scene representation.
%
%
%
\subsubsection{Metrics}
To measure image synthesis quality, we first mask all results using the eroded ground truth foreground masks since even ground truth masks still contain segmentation errors.
Otherwise falsely classified background pixels, that are however correctly recovered by the respective methods would erroneously lead to high errors.
Then, we evaluate the peak signal-to-noise ratio (PSNR).
However, this metric does not reflect the visual perception of humans, i.e., blurry results can have a low error although they appear very unrealistic to the human eye~\cite{8578166}.
Hence, we also report the learned perceptual image patch similarity (LPIPS)~\cite{8578166}, and the Fréchet inception distance (FID)~\cite{heusel17} metrics, which are human perception-based metrics.
In contrast to our approach, other methods cannot generate results at 4K resolution ($4112 \times 3008$) in a reasonable time, we evaluate all metrics on the downsampled versions ($1285 \times 940$) if not specified otherwise.
To measure geometry quality, we report the Chamfer and Hausdorff distance between the pseudo ground truth and the reconstructed results.
%
%
\subsubsection{Image Synthesis Accuracy}
\begin{figure*}
	\begin{center}
		\includegraphics[width=\linewidth]{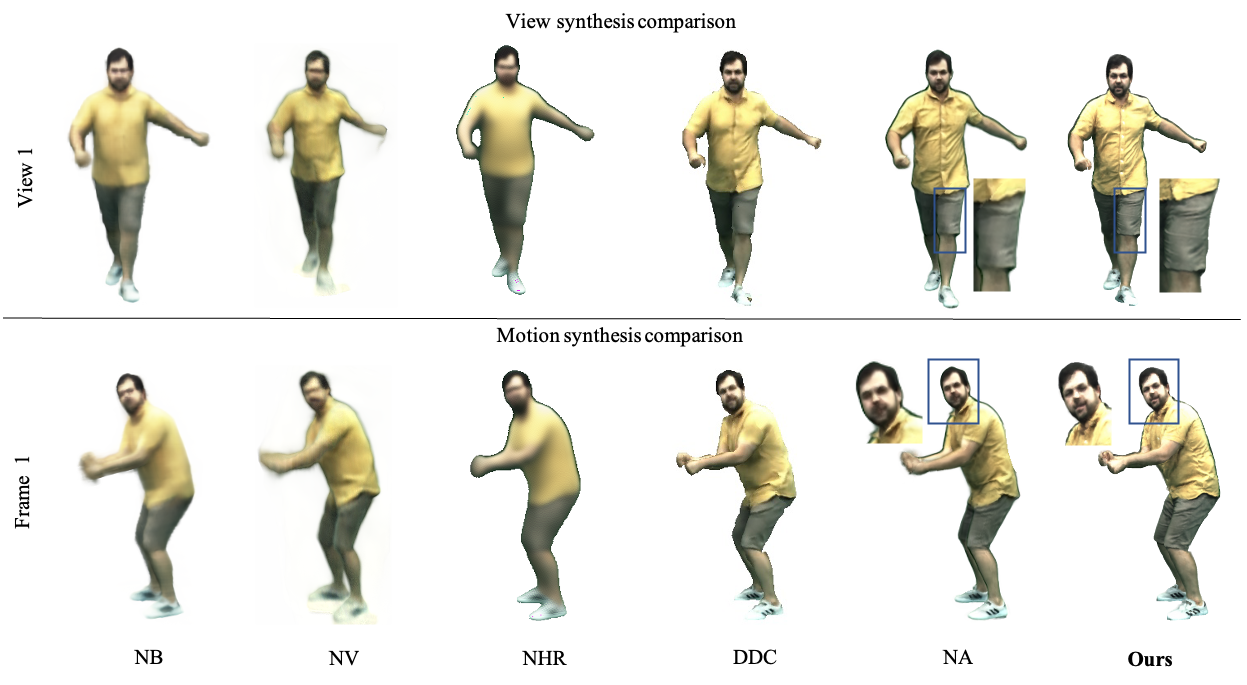}
	\end{center}
	\caption
	{
    	Here, we visually compare our approach in terms of the image synthesis quality to two recent neural human rendering methods, Neural Actor~\cite{liu2021neural}, and Deep Dynamic Characters~\cite{habermann21}.
    	Note that our approach renders sharper images and smaller details in the image can be much better recovered by our method compared to the previous works.
	}
	\label{fig:comparison_synthesis}
\end{figure*}

First, we evaluate the image synthesis quality of our approach and compare it to previous works.
In Figure~\ref{fig:comparison_synthesis}, we show a qualitative comparison to previous works. 
For NB~\cite{peng2020neural}, NV~\cite{lombardi2019neural}, and NHR~\cite{wu2020multi}, the results are very blurry and contain obvious visual artifacts. 
Compared to their original results, we found that when using the larger and more challenging \textit{DynaCap} training dataset, which also contains more variations, the quality of these methods significantly degrades.
Thus, these methods seem to be inherently limited to shorter sequences and exhibit limited generalization ability in terms of unseen poses and views.
The results of DDC~\cite{habermann21} are less blurry compared to the aforementioned methods, but high frequency wrinkles are still not recovered well.
In contrast, NA~\cite{liu2021neural} captures such wrinkles, but as mentioned earlier this work can only handle tight types of clothing.
In contrast to that, our method is able to synthesize arbitrary types of apparel and also produces the sharpest and most detailed results.
%
%
%
%
%
%
%
\par
Table~\ref{tab:view_error} also quantitatively confirms that our method achieves the best view synthesis results in terms of perceptual metrics.
We provide the numbers for our method when we did \textit{not} use 4K videos during training (referred to as \textit{Ours w/o 4k}). 
Importantly, other methods cannot be trained on 4K video data in a reasonable amount of time while our method design allows training on such data in general.
Notably, we outperform other approaches in terms of LPIPS and FID by 27.2\% and 73.4\%.
The difference in PSNR however is less prominent since this metric is less sensitive to blurry results and, thus, even if results are more blurry the PSNR can be higher~\cite{8578166}.
This explains why DDC has a slightly better score even though our results are significantly more plausible.
As stated in Section~\ref{sec:motion_texture}, we found that the motion to appearance mapping is an one-to-many mapping~\cite{bagautdinov21, liu2021neural}.
While others~\cite{peng2020neural, lombardi2019neural, wu2020multi, habermann21, su2021anerf} ignore this, we explicitly account for it by leveraging a discriminator during the training of the texture network.
Thus, our generated textures are sharp and plausible, but the exact wrinkles might vary from the ground truth since there is no unique mapping.
However, they will not fully align with the ground truth resulting in a lower PSNR compared to a blurred result.
To confirm this, we also evaluate our method when using the ground truth texture maps instead of the synthesized texture map such that wrinkles in the texture align with the real ones, referred to as \textit{Ours w/ GT textures}.
The PSNR then clearly outperforms previous works.
%
%
%
\begin{table}
\begin{center}
\caption
{
View synthesis error of the \textit{D2} sequence.
%
%
Note that we achieve by far the highest scores for the perception-based metrics and also in terms of PSNR our method performs better than the previous state-of-the-art methods.
}
\label{tab:view_error}	
\begin{tabular}{|c|c|c|c|}
    \hline
    \multicolumn{4}{|c|}{\textit{View synthesis error} on \textit{D2}} \\
    \hline
    \textbf{Method}  & \textbf{PSNR}$\uparrow$  & \textbf{LPIPS}$\downarrow$ ($\times 1000$)  & \textbf{FID}$\downarrow$ \\
    \hline
    NB	                        &   29.94                &  42.15         &   109.98            \\ 
    NV	                        &   25.49                &  85.69         &   123.19            \\ 
    NHR	                        &   28.39                &  46.07         &   116.59            \\ 
    DDC	                        &   \textbf{32.96}                &  20.07         &   27.73            \\ 
    NA	                        &   30.21                &  23.60         &   18.56            \\ 
    A-NeRF	                    &   29.54               &      35.27           &     86.90          \\ 
    \textbf{Ours w/o 4k}	                &   31.00                &  \textbf{14.61}        &   \textbf{4.93}          \\
    \hline
    Ours w/ GT textures	    &   33.96                &  10.87         &   5.15            \\
    \hline
\end{tabular}
\end{center}
\end{table}
%
%
%
%
\par
The same tendency can be observed when comparing to other works in terms of motion synthesis (see Table~\ref{tab:motion_error}).
Again, our method achieves the best perceptual results due to the high quality synthesis of our approach.
In terms of PSNR, some methods achieve a higher score although results are notably very blurred and/or not photo-real.
The reason is the same as before, which also for this setting can be confirmed when evaluating the PSNR value for \textit{Ours w/ GT textures}.
However, as confirmed by our qualitative results and the supplemental video, our method clearly outperforms previous works in terms of perceived image quality  and photorealism.
%
%
%
\begin{table}
\begin{center}
\caption
{
We also evaluate the motion synthesis quality of our approach and compare it previous methods.
Again, we outperform other works in terms of the perception-based metrics and are comparable to earlier works in terms of PSNR.
}
\label{tab:motion_error}	
\begin{tabular}{|c|c|c|c|}
    \hline
    \multicolumn{4}{|c|}{\textit{Motion synthesis error} on \textit{D2}} \\
    \hline
    \textbf{Method} & \textbf{PSNR}$\uparrow$  & \textbf{LPIPS}$\downarrow$ ($\times 1000$) & \textbf{FID}$\downarrow$ \\
    \hline
    NB	                        &    \textbf{29.37}           &    43.99           &      115.70         \\ 
    NV	                        &    23.32           &    98.35             &    139.82           \\ 
    NHR	                        &    28.08           &    47.65           &      122.60         \\ 
    DDC	                        &    28.05           &    30.43             &    38.37           \\ 
    NA	                        &    28.43           &    28.33             &    24.50           \\ 
    A-NeRF	            &    28.42            &   38.74           &     95.56          \\ 
    \textbf{Ours w/o 4k}	                &    27.69          &    \textbf{24.0}             &     \textbf{9.25}         \\
    \hline
    Ours w/ GT textures	    &   31.32                &  15.84         &   9.34           \\ 
    \hline
\end{tabular}
\end{center}
\end{table}
%
%
%
%
\subsubsection{Geometry Deformation Accuracy}
\begin{figure}
	\begin{center}
		\includegraphics[width=\linewidth]{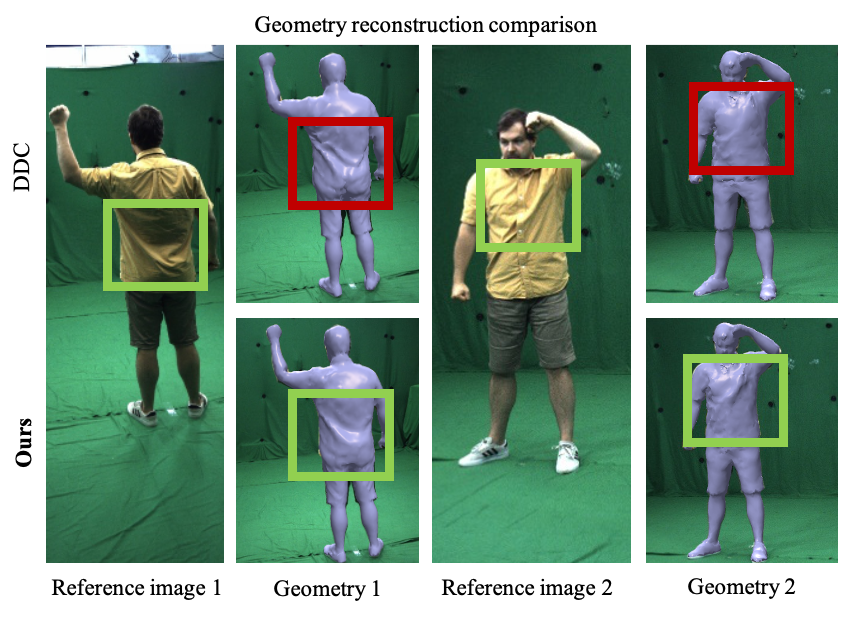}
	\end{center}
	\caption
	{
		We compare our surface deformation quality to previous work~\cite{habermann21}.
		Due to our NeRF-guided supervision, geometric details can be better tracked compared to DDC, which solely supervises the geometry in image space.
	}
	\label{fig:comparison_geometry}
\end{figure}

In Figure~\ref{fig:comparison_geometry} and Table~\ref{tab:reconstruction_error}, we qualitatively and quantitatively evaluate the surface deformation accuracy of our approach and compare it to DDC~\cite{habermann21} and NA~\cite{liu2021neural}.
For NA, we used Marching Cubes to retrieve per-frame reconstructions.
The recovered geometries contain a lot of noise since the density field is not regularized, thus, resulting in worse performance.
DDC is the only previous work that also tracks the space-time coherent deformation of the template.
One can see that our method has an overall lower error in terms of surface quality, which is due to our NeRF-guided supervision.
Further, new wrinkle patterns, which appear while the actor is performing different motions, cannot be tracked well by DDC, i.e., the wrinkles in the geometry often do not match the ones in the images as indicated by the red boxes in Figure~\ref{fig:comparison_geometry}.
In contrast, we demonstrate that our NeRF guidance helps the deforming geometry to recover these details.
%
%
%
%
%
%
%
%
\begin{table}
	\begin{center}
    	\caption
    	{
    	    Here, we evaluate the 3D error on the training skeletal motion of the \textit{D2} sequence.
    	    Note that the proposed method outperforms previous work in terms of tracking the surface deformation.
            We further evaluate the 3D error on the test skeletal motion of the challenging \textit{D2} sequence.
    	    Note that the proposed method also outperforms previous work in terms of deformation synthesis.
    	    The error is reported in $cm$.
    	}
    	\label{tab:reconstruction_error}	
		\begin{tabular}{|c|c|c|}
			\hline
			\multicolumn{3}{|c|}{\textit{Reconstrution Error} on \textit{D2}} \\
			\hline
			\textbf{Method}                                 		& \textbf{Chamfer} $\downarrow$   	            & \textbf{Hausdorff} $\downarrow$        \\
			\hline
		    NA							            & 2.21		                                        & 2.75                            \\ 
		    DDC							            & 1.2686		                                        & 1.1922                               \\ 
			\textbf{Ours w/o 4k}							                & \textbf{1.0071}		                                        & \textbf{0.8872}                       \\ 
			\hline
			\hline
			\multicolumn{3}{|c|}{\textit{Deformation generation error} on \textit{D2}} \\
			\hline
			\textbf{Method}                                 		& \textbf{Chamfer} $\downarrow$   	            & \textbf{Hausdorff} $\downarrow$     \\
			\hline
		    NA							            & 2.29		                                        & 2.82                            \\ 
		    DDC							            & 1.43		                                        & 1.38                           \\ 
			\textbf{Ours w/o 4k}							                & \textbf{1.24}		                                        & \textbf{1.15}                        \\ 
			\hline
		\end{tabular}
	\end{center}
\end{table}
%
%
%
%
%
\subsection{Ablation Studies} \label{sec:ablation}
%
%
%
\begin{figure}
	\begin{center}
		\includegraphics[width=\linewidth]{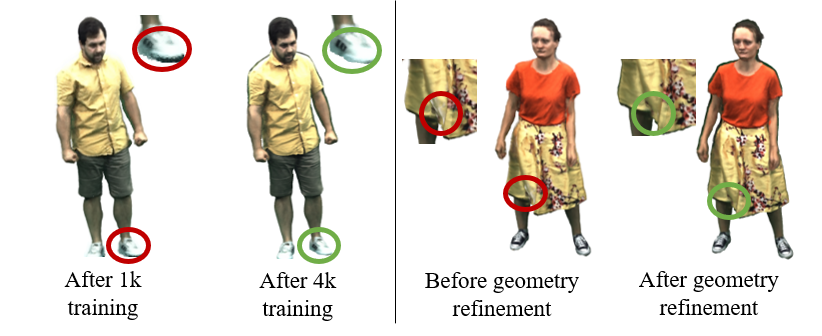}
	\end{center}
	\caption
	{
		We evaluate the effect of the 4k ($4112 \times 3008$) image supervision with using only 1k ($1285 \times 940$). 
		The 4k resolution shows a superior quality compared to only training on 1k images.
		Further, the influence of the geometry refinement also greatly improves the synthesis quality.
	}
	\label{fig:ablation_synthesis}
\end{figure}
%
%
%
%
\begin{table}
\begin{center}
\caption
{
Ablation study.
Here, we evaluate the influence of our proposed mesh refinement and the influence of training on 4k resolution instead of 1k resolution. 
Note that the refinement greatly improves the result across all metrics.
When testing on 4k resolution, one can see that also the training on 4k improves all metrics.
}
\label{tab:ablation}	
\begin{tabular}{|c|c|c|c|}
    \hline
    \multicolumn{4}{|c|}{\textit{Motion synthesis error} on \textit{D2}} \\
    \hline
    \textbf{Method} & \textbf{PSNR}$\uparrow$  & \textbf{LPIPS}$\downarrow$ ($\times 1000$) & \textbf{FID}$\downarrow$ \\
    \hline
    Ours w/o ref. and 4k        &    27.49           &    25.64             &    12.90           \\
    \textbf{Ours w/o 4k}	                &    \textbf{27.69}          &    \textbf{24.0}             &     \textbf{9.25}         \\
    \hline
    Ours w/o 4k	($4112 \times 3008$)                &    26.02         &    50.38           &   16.21        \\
    \textbf{Ours ($\mathbf{4112 \times 3008}$)}	            &    \textbf{26.04}        &    \textbf{49.81}           &    \textbf{15.13}       \\
    \hline
    \hline
    \multicolumn{4}{|c|}{\textit{View synthesis error} on \textit{D2}} \\
    \hline
    \textbf{Method}  & \textbf{PSNR}$\uparrow$  & \textbf{LPIPS}$\downarrow$ ($\times 1000$)  & \textbf{FID}$\downarrow$ \\
    \hline
    Ours w/o ref. and 4k	&   29.89                &  17.63         &   8.54            \\
    \textbf{Ours w/o 4k}	                &   \textbf{31.00}                &  \textbf{14.61}        &   \textbf{4.93}          \\
    \hline
    Ours w/o 4k ($4112 \times 3008$)	        &   29.11               &  46.11         &   12.82        \\
    \textbf{Ours ($\mathbf{4112 \times 3008}$)}	            &   \textbf{29.32}            &    \textbf{43.87}        &   \textbf{11.26}       \\ 
    \hline
\end{tabular}
\end{center}
\end{table}
%
%
%
%
\subsubsection{Deformable Mesh Guidance}
First, we evaluate the design choice of uniting an explicit and \textit{deformable} mesh representation with a neural radiance field.
To this end, we leverage the pre-trained deformation networks of DDC~\cite{habermann21} to obtain the deformed geometry and apply our mesh-guided radiance field (Section~\ref{sec:meshguided}); this method is referred to as \textit{Ours w/o ref. and 4k} in Table~\ref{tab:ablation}.
Note that we show a consistent improvement in terms of perceptual metrics over NA~\cite{liu2021neural} and DDC~\cite{habermann21}.
NA can be considered as a method that leverages a explicit \textit{piece-wise rigid} mesh to guide a neural radiance field. 
In contrast, DDC explicitly accounts for non-rigid mesh deformations, but it does not leverage a NeRF representation.
Thus, this baseline comparison clearly shows the advantage of uniting a \textit{deformable} mesh representation with a neural radiance field.
%
%
\subsubsection{Geometry Refinement}
Next, we evaluate the influence of the NeRF-guided geometry refinement (Section~\ref{sec:nerfguided}).
In Figure~\ref{fig:ablation_synthesis}, we show how the better geometry tracking helps to achieve a higher synthesis quality as wrinkle patterns appear sharper.
This is also quantitatively confirmed in Table~\ref{tab:ablation} where the result with our proposed refinement (\textit{Ours w/o 4k}) is consistently better than our result without refinement (\textit{Ours w/o ref. and 4k}).
In terms of geometry error, we found that compared to the baseline (DDC), the NeRF-guided loss also helps to recover geometry that is closer to the ground truth (see Table~\ref{tab:reconstruction_error}).
Thus, for both tasks, image synthesis and surface recovery, the proposed NeRF-guided geometry refinement improves the results.
Importantly, the joint consideration of deformation tracking and synthesis, for the first time, allows to achieve such photo-real quality for \textit{loose types of apparel}. 
%
%
\subsubsection{4K Supervision}
We further study the influence of the 4k image resolution in Figure~\ref{fig:ablation_synthesis}.
Again, one can see that more details in the image can be preserved when 4k images are used for training, i.e., the black stripes on the shoe are sharper.
This is also quantitatively confirmed in Table~\ref{tab:ablation} where we evaluate the metrics on the \textit{higher resolution} ($4112 \times 3008$) images instead of the downsampled ones. 
The error can be further reduced by using the 4K supervision and that training on such data is only possible in an acceptable time due to our more efficient mesh-based sampling and feature attachment strategy.
In fact, the original NeRF architecture (using the fine MLP as well) requires $64+128$ samples per ray whereas our proposed architecture and sampling only requires 32 samples.
This reduces our training time to 10 days compared to 29 days when using the original architecture.
%
%

\subsection{Applications}\label{sec:application}
%
%
%
\begin{figure}
	\begin{center}
		\includegraphics[width=\linewidth]{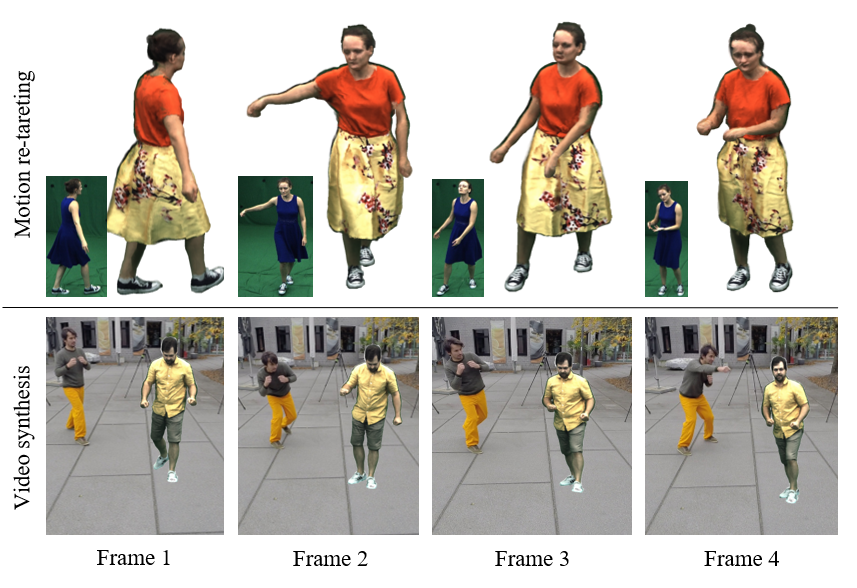}
	\end{center}
	\caption
	{
	Our method also enables exciting applications. 
	Here, we show a motion re-targeting result of our method where we apply the motion of the actor with the blue dress on the actor with the yellow skirt.
	Moreover, our method can be used for video synthesis. 
	To this end, we overlay our actors onto a dynamic video.
	For both applications, our method achieves photorealistic results.
	}
	\label{fig:applications}
\end{figure}
%
%
%
In addition to view and motion synthesis, our method can also enable other exciting applications such as motion re-tarteting. 
Figure~\ref{fig:applications} shows such an application setting where the actor with the blue dress is driving the actor with the yellow skirt.
Note that the resulting images are very sharp and even small-scale wrinkles can be synthesized. 
Further, our method enables video synthesis where we augment an existing video with our virtual and photo-realistic characters.
%
%
\section{Limitations and Future Work} \label{sec:limitations}
Although our method achieves state-of-the-art results in terms of view and motion synthesis, it still has some limitations, which require future research in this direction.
First, we currently do not capture and model hand gestures and facial expressions.
This results sometimes in blurry results in these regions and these body parts can also not be explicitly controlled by the user.
In the future, we plan to explore this direction to provide a fully controllable digital avatar.
Moreover, our method does not model the incoming light independently from the reflectance properties of the surface.
This comes with the limitation that the light is "baked-into" the appearance and novel lighting conditions cannot be synthesized.
Here, a more explicit decomposition of light and material could potentially solve the problem.
We rely on the motion tracking quality of our marker-less motion capture system and failures in the tracking can lead to artifacts in our results.
To overcome this, one could jointly refine the skeletal pose by backpropagating the dense color losses through the entire neural rendering pipeline.
Moreover, some artifacts around the boundary between the actor and the background arise from wrong ground truth segmentations during training.
Here, we plan to investigate whether the pixel-wise classification can be jointly estimated during training.
Last, even though our method is significantly more efficient than the baseline and capable of rendering 4K images within a few seconds, the training time could still be improved and the inference time is not yet real time.
In the future, we would like to investigate alternative (potentially more lightweight) network designs~\cite{yu_and_fridovichkeil2021plenoxels,Garbin_2021_ICCV, Chan2022} and further explore the promising idea of hybrid representations.
%

%
%
\section{Conclusion} \label{sec:conclusion}
We proposed \hdh a method for view and motion synthesis of digital human characters from multi-view videos.
Our method solely takes a skeletal motion and a camera pose and produces high resolution images and videos of an unprecedented quality.
At the technical core, we propose to jointly learn the surface deformation of the human and the appearance in form of a neural radiance field.
We showed that this has a synergy effect and the combination of both scene representations improves each other.
Our results demonstrate that our method is a clear step forward towards more photo-realistic and higher resolution digital avatars, which will be an important part for the upcoming era of AR and VR.
We also believe that our work can be a solid basis for future research in this direction, which potentially tackles the challenges of real-time compute, relighting, and face and hand gesture synthesis.

\begin{acks}
All data captures and evaluations were performed at MPII by MPII.
The authors from MPII were supported by the ERC Consolidator Grant 4DRepLy (770784), the Deutsche Forschungsgemeinschaft (Project Nr. 409792180, Emmy Noether Programme, project: Real Virtual Humans) and Lise Meitner Postdoctoral Fellowship.
Gerard Pons-Moll was supported by German Federal Ministry of Education and Research (BMBF): Tuebingen AI Center, FKZ: 01IS18039A.
\end{acks}


\bibliographystyle{ACM-Reference-Format}
\bibliography{mybib.bib}

\appendix
%
%
\section{Implementation}\label{sec:implementation}
In the following, we provide more details regarding the individual network architectures and training stages.
As mentioned earlier, we pre-train $f_\mathrm{eg}$ and $f_\mathrm{delta}$ as described in \cite{habermann21}.
%
%
\par \textbf{Details for $f_\mathrm{tex}$.}
For $f_\mathrm{tex}$, we use the vid2vid~\cite{wang2018vid2vid} network with the default setting to predict texture maps at a resolution of $1024 \times 1024$ from normal maps in $1024 \times 1024$. 
We trained vid2vid for about 30k iterations. 
%
%
\par \textbf{Details for $f_\mathrm{enc}$ and $f_\mathrm{nf}$.}
For the texture encoder $f_\mathrm{enc}$, we leverage the Tensorflow version of the pix2pix architecture~\cite{pix2pix2017}. 
We added two downsampling and two upsampling layers and adjusted the feature channel size in order to encode and decode images at a resolution of $1024 \times 1024$.
This network is trained jointly with $f_\mathrm{nf}$.
Important to note is that we train the encoder with the ground truth texture maps while we use the predicted texture maps only at test time.
For $f_\mathrm{nf}$, we leverage the architecture proposed by Mildenhall et al.~\shortcite{mildenhall2020nerf} with some changes.
We changed the network depth to 16 for both the density predicton module as well as the view-dependent branch.
Further, we changed the number of activations of the density module and the view-dependent branch to 128 and 64, respectively.
After every 4 fully connected layers we employ a skip connection.
We train the network with the Adam optimizer~\cite{kingma14} and employ a learning rate of $0.0005$.
Each training batch contains 1024 samples.
We first train on downsampled images ($1285\times940$) for $2$ million iterations.
Here, one batch contains 8 different camera views each sampling 128 rays for a randomly chosen frame.
Then, the mesh is refined using the proposed NeRF-guided supervision, before $f_\mathrm{enc}$ and $f_\mathrm{nf}$ are once more trained for another $500k$ iterations using the refined geometry.
Last, $f_\mathrm{enc}$ and $f_\mathrm{nf}$ are refined on the 4K images ($4112 \times 3008$) for $1.5$ million iterations.
Here, we only sample 1 camera view and frame per batch.
%
%
\par \textbf{Details for $f_\mathrm{delta}$.}
For $f_\mathrm{delta}$, we adopt the architecture proposed by Habermann et al.~\shortcite{habermann21}.
We refine the pre-trained network for $360k$ iterations using the novel NeRF-guided loss.
We weight the individual loss terms with $\lambda_\mathrm{pc}=5000.0$, $\lambda_\mathrm{sil}=50.0$, $\lambda_\mathrm{laplacian}=4000.0$, and $\lambda_\mathrm{edge}=0.075$.
%
%

%
%
\section{Training Procedure} \label{sec:training}
We first train $f_\mathrm{enc}$ and $f_\mathrm{nf}$.
Once trained, we refine the deformation network using the NeRF-guided loss.
Then, we use the refined geometry to train $f_\mathrm{enc}$ and $f_\mathrm{nf}$ once more, first on the lower resolution images and later on the $4k$ images.
This procedure could potentially be iterated multiple times, but we found that in practice a single iteration is sufficient.
All our experiments are performed on a machine with 2 to 4 NVIDIA Quadro RTX 8000 48Gb graphics cards and an AMD EPYC 7502P 32-Core processing unit.
For each subject, the training of the NeRF MLP and the texture encoder leverages 2 GPUs and takes about 9 days.
Refining the surface deformation network also leverages 2 GPUs and takes 1.5 days.
The training of the texture translation network requires 4 GPUs and takes 7 days.
However, this step can be trained in parallel since the NeRF MLP and the texture encoder leverage the ground truth textures during training.
Thus, the total training time is around 10 days.
At test time, our approach requires around 12 seconds to generate a 4K image using a single GPU.


\end{document}